\documentclass[10pt,twocolumn,letterpaper]{article}

\usepackage[pagenumbers]{cvpr}      % To produce the REVIEW version
\usepackage{graphicx}
\usepackage{amsmath}
\usepackage{amssymb}

\usepackage{multirow}
\usepackage{booktabs}
\usepackage{tabulary}
\usepackage[dvipsnames]{xcolor}
\usepackage{xspace}\xspace
\usepackage{adjustbox}
\usepackage{array}

\usepackage[pagebackref,breaklinks,colorlinks]{hyperref}

\newcolumntype{x}[1]{>{\centering\arraybackslash}p{#1pt}}
\newcommand{\cgaphl}[2]{
\fontsize{8pt}{1em}\selectfont{${#1}$\textbf{#2}}
}
\newcommand{\cgaphlp}[2]{
\fontsize{6pt}{-.5em}\selectfont{(${#1}$\textbf{#2})}
}
\newlength\savewidth\newcommand\shline{\noalign{\global\savewidth\arrayrulewidth
  \global\arrayrulewidth 1pt}\hline\noalign{\global\arrayrulewidth\savewidth}}
\newcommand{\tablestyle}[2]{\setlength{\tabcolsep}{#1}\renewcommand{\arraystretch}{#2}\centering\footnotesize}
\makeatletter\renewcommand\paragraph{\@startsection{paragraph}{4}{\z@}
  {.5em \@plus1ex \@minus.2ex}{-.5em}{\normalfont\normalsize\bfseries}}\makeatother

\usepackage[capitalize]{cleveref}
\crefname{section}{Sec.}{Secs.}
\Crefname{section}{Section}{Sections}
\Crefname{table}{Table}{Tables}
\crefname{table}{Tab.}{Tabs.}

\def\confName{CVPR}
\def\confYear{2022}

\begin{document}

% \title{MixFormer: A Vision Transformer with Comprehensive Feature Mixing}
\title{MixFormer: Mixing Features across Windows and Dimensions}

\author{
Qiang Chen$^1$\thanks{Equal Contribution.} , Qiman Wu$^{1*}$, Jian Wang$^{1*}$, Qinghao Hu$^2$\thanks{Corresponding author.} , Tao Hu$^1$\\ Errui Ding$^1$, Jian Cheng$^2$, Jingdong Wang$^1$\\
$^1$Baidu VIS\\
$^2$NLPR, Institute of Automation, Chinese Academy of Sciences\\
{\tt\small \{chenqiang13,wuqiman,wangjian33,hutao06,dingerrui,wangjingdong\}@baidu.com}\\
{\tt\small huqinghao2014@ia.ac.cn, jcheng@nlpr.ia.ac.cn}
}
\maketitle

%%%%%%%%% ABSTRACT
\begin{abstract}
While local-window self-attention performs notably in vision tasks, it suffers from limited receptive field and weak modeling capability issues. This is mainly because it performs self-attention within non-overlapped windows and shares weights on the channel dimension. We propose MixFormer to find a solution. First, we combine local-window self-attention with depth-wise convolution in a parallel design, modeling cross-window connections to enlarge the receptive fields. Second, we propose bi-directional interactions across branches to provide complementary clues in the channel and spatial dimensions. These two designs are integrated to achieve efficient feature mixing among windows and dimensions. Our MixFormer provides competitive results on image classification with EfficientNet and shows better results than RegNet and Swin Transformer. Performance in downstream tasks outperforms its alternatives by significant margins with less computational costs in 5 dense prediction tasks on MS COCO, ADE20k, and LVIS. Code is available at  \url{https://github.com/PaddlePaddle/PaddleClas}.
\end{abstract}

%%%%%%%%% BODY TEXT
%-------------------------------------------------------------------------
\section{Introduction}
\label{sec:intro}
%-------------------------------------------------------------------------
The success of Vision Transformer (ViT)~\cite{dosovitskiy2020image,touvron2021training} in image classification~\cite{deng2009imagenet} validates the potential to apply Transformer~\cite{vaswani2017attention} to vision tasks. Challenges remain for downstream tasks, especially the inefficiency in high-resolution vision tasks and the ineffectiveness in capturing local relations. One possible solution is to use local-window self-attention. It performs self-attention within non-overlapped windows and shares weights on the channel dimension. Although this process improves efficiency, it poses the issues of limited receptive field and weak modeling capability.
%##################################################################################################
\begin{figure}
\centering
\includegraphics[width=0.35\textwidth]{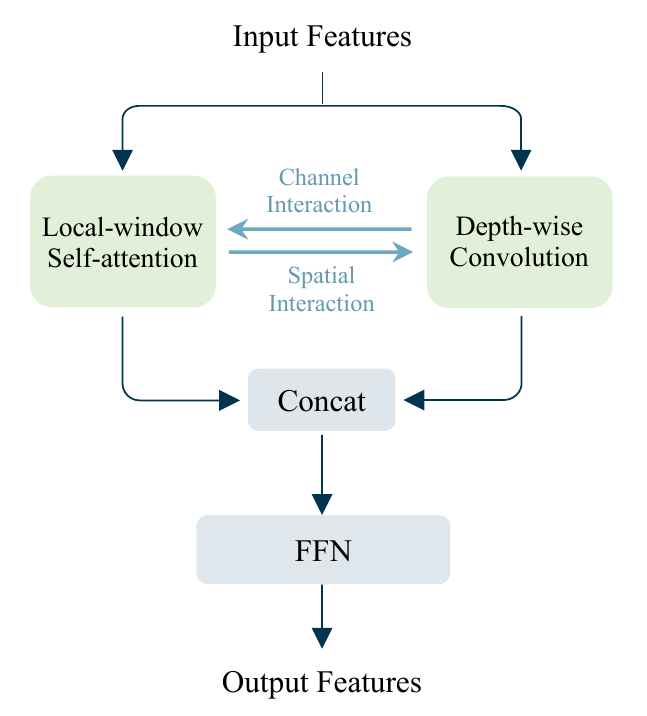}
\caption{\textbf{The Mixing Block.} We combine local-window self-attention with depth-wise convolution in a parallel design. The captured relations within and across windows in parallel branches are concatenated and sent to the Feed-Forward Network (FFN) for output features. In the figure, the blue arrows marked with {\em Channel Interaction} and {\em Spatial Interaction} are the proposed bi-directional interactions, which provide complementary clues for better representation learning in both branches. Other details in the block, such as module design, normalization layers, and shortcuts, are omitted for a neat presentation.}
\label{fig:1}\vspace{-5mm}
\end{figure}
%##################################################################################################

A common approach to expand receptive field is to create cross-window connections. Windows are connected by shifting~\cite{liu2021swin}, expanding~\cite{vaswani2021scaling,yang2021focal}, or shuffling~\cite{huang2021shuffle} operations. Convolution layers are also employed as they capture natural local relations. Researches~\cite{huang2021shuffle,YuanFHLZCW21} combine local-window self-attention with depth-wise convolution base on this and provide promising results. Still, the operations capture intra-window and cross-window relations in successive steps, leaving these two types of relations less interweaved. Besides, neglect of modeling weakness in these attempts hinders further advances in feature representation learning.

We propose Mixing Block to address both these issues (Figure~\ref{fig:1}). First, we combine local-window self-attention with depth-wise convolution, but in a parallel way. The parallel design enlarges the receptive fields by modeling intra-window and cross-window relations simultaneously. Second, we introduce bi-directional interactions across branches(illustrated as blue arrows in Figure~\ref{fig:1}). The interactions offset the limits caused by the weight sharing mechanism\footnote{Local-window self-attention shares weights on the channel dimension while depth-wise convolution shares weights on the spatial one~\cite{han2021demystifying}. From the weight sharing perspective, sharing weights results in limited modeling ability in the correspond dimension.}, and enhance the modeling ability in channel and spatial dimensions by providing complementary clues for local-window self-attention and depth-wise convolution respectively. The above designs are integrated to achieve complementary feature mixing across windows and dimensions.

We present MixFormer to verify the block’s efficiency and effectiveness. A series of MixFormers with computational complexity ranging from 0.7G (B1) to 3.6G (B4) are built to perform distinguished in multiple vision tasks, including image classification, object detection, instance segmentation, semantic segmentation, etc. On ImageNet-1K~\cite{deng2009imagenet}, we achieve competitive results with EfficientNet~\cite{tan2019efficientnet}, surpassing RegNet~\cite{radosavovic2020designing} and Swin Transformer~\cite{liu2021swin} by a large margin. MixFormer markedly outperforms its alternatives in 5 dense prediction tasks with lower computational costs. With Mask R-CNN~\cite{he2017mask}($1\times$) on MS COCO~\cite{lin2014microsoft}, MixFormer-B4 shows a boost of 2.9 box mAP and 2.1 mask mAP on Swin-T~\cite{liu2021swin} while requiring less computational cost. Substituting the backbone in UperNet~\cite{xiao2018unified}, MixFormer-B4 delivers a 2.2 mIoU gain over Swin-T~\cite{liu2021swin} on ADE20k~\cite{zhou2019semantic}. Plus, MixFormer is effective in keypoint detection~\cite{lin2014microsoft} and long-tail instance segmentation~\cite{gupta2019lvis}. In brief, our MixFormer achieves state-of-the-art performance on multiple vision tasks as an efficient general-purpose vision transformer. 
% Our contributions are summarized as follows.

%-------------------------------------------------------------------------
\section{Related Works}
\label{sec:relatedworks}
%-------------------------------------------------------------------------
%##################################################################################################
\begin{table*}[t]
\tablestyle{1.5pt}{1.2}
\begin{tabular}{x{65}x{65}x{65}x{65}x{65}}
\shline
& Attention & W-Attention & Conv   & DwConv \\
\hline
Sharing Weights & Channel Dim & Channel Dim & Spatial Dim &  Spatial Dim\\
FLOPs  & $2NCH^2W^2$ & $\pmb{2NCHWK^2}$ & $NC^2HWK^2$ &  $\pmb{NCHWK^2}$ \\
\shline
\end{tabular}
\caption{\textbf{Sharing Weights Dimensions and FLOPs.} We provide comparison among four operations: global self-attention(Attention), local window self-attention(W-Attention), convolution(Conv) and depth-wise convolution(DwConv). In the table, we provide the dimension of weight sharing for all components in the first row.
Besides, the FLOPs is calculated with a $N \times C \times H \times W$ input and a output with the same shape. The K in the table represents the window size in local-window self-attention or convolution. Note that the Attention operator adopts a window size of $H\times W$ as it models global dependencies in the spatial dimension.}
\label{tab:2}\vspace{-4mm}
\end{table*}
%##################################################################################################
%##################################################################################################
\begin{figure*}
\centering
\includegraphics[width=0.65\textwidth]{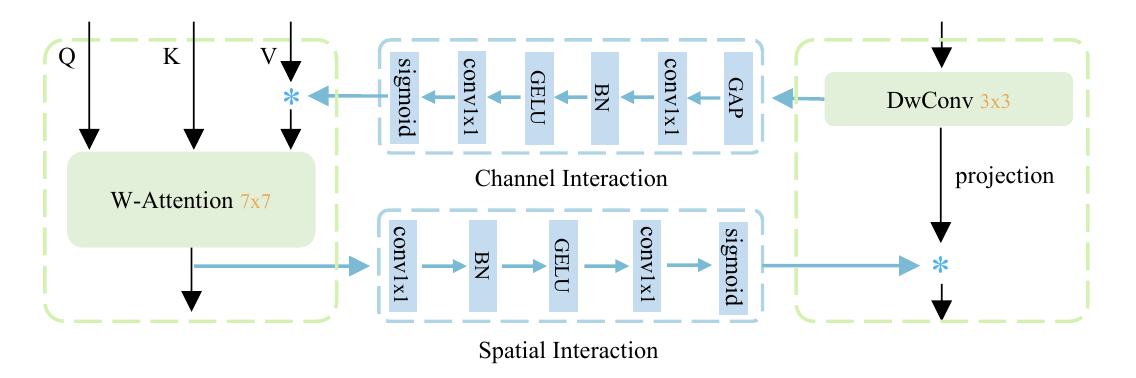}
\caption{\textbf{Detailed design of the Bi-directional Interactions.} The channel/spatial interaction provides channel/spatial context extracted by depth-wise convolution/local-window self-attention to the other path.}
\label{fig:2}\vspace{-3mm}
\end{figure*}
%##################################################################################################
\paragraph{Vision Transformers.} The success of the pioneering work, ViT~\cite{dosovitskiy2020image,touvron2021training}, shows great potentials to apply transformer to the computer vision community. After that, various methods~\cite{touvron2021training,han2021transformer,yuan2021tokens,jiang2021token,zhou2021deepvit,bai2021visual} are proposed to improve the performance of vision transformers, demonstrating competitive results on the image classification task. As the self-attention ~\cite{vaswani2017attention} is different from the convolution in nature: self-attention models long-range dependencies while convolution captures relations in local windows, there are also works aiming for integrating convolution and vision transformer. Works like PVT~\cite{wang2021pyramid} and CvT~\cite{wu2021cvt} insert spatial reduction or convolution  before global self-attention, yielding the merits of self-attention and convolution. 
\paragraph{Window-based Vision Transformers.} Although global Vision Transformer shows its success on image classification; challenges remain for downstream tasks. For high-resolution vision tasks, the computation cost of the Vision Transformer is quadratic to image size, making it unaffordable for real-world applications. Recently, researchers have proposed plenty of methods ~\cite{yan2021contnet,wang2021pyramid,wu2021cvt,liu2021swin,dong2021cswin,chu2021twins} to make vision transformers become general-purpose backbones as ConvNets~\cite{he2016deep,xie2017aggregated,howard2017mobilenets}. Among them, Window-based Vision Transformer~\cite{chu2021twins, liu2021swin, huang2021shuffle} adopts the local window attention mechanism, making its computational complexity increase linearly to image size.

\paragraph{Receptive Fields.}
Receptive fields are important for the downstream vision tasks. However, Window-based Vision Transformer computes self-attention within non-overlapping local windows, which limits the receptive fields in local windows. To solve the problem, researchers propose to use shifting~\cite{liu2021swin}, expanding~\cite{vaswani2021scaling,yang2021focal}, or shuffling~\cite{huang2021shuffle} operations to connect nearby windows.There are also works \cite{YuanFHLZCW21,huang2021shuffle} using convolutions to enlarge the receptive fields efficiently. Convolution layers are used to create connections because they capture local relations in nature. We combine local-window self-attention and depth-wise convolution in our block design.

\paragraph{Dynamic Mechanism.}
Dynamic networks here ~\cite{hu2018squeeze,woo2018cbam,li2019selective,jaderberg2015spatial,dai2017deformable,vaswani2017attention} refer to networks whose parts of weights or paths are data-dependent. Generally speaking, the dynamic network achieves higher performance than its static alternative as it is more flexible in modeling relations. In ConvNets, the dynamic mechanism is widely used to better extract customized features given different inputs. There are various types of dynamic networks that focus on the channel~\cite{hu2018squeeze,li2019selective} and the spatial dimension~\cite{jaderberg2015spatial,dai2017deformable,woo2018cbam}. These works promote many tasks to new state-of-the-art. For Transformer~\cite{vaswani2017attention}, the self-attention module is a dynamic component, which generates attention maps based on the inputs. In this paper, we also adopt the dynamic mechanism in the network design, while our application is based on the finding that the two efficient components share their weights on different dimensions~\cite{han2021demystifying}. To construct a powerful block while maintaining efficiency, we introduce dynamic interactions across two branches, which are light-weighted and improve the modeling ability in both channel and spatial dimensions. 
%-------------------------------------------------------------------------
\section{Method}
\label{sec:method}
\subsection{The Mixing Block}
Our Mixing Block (Figure~\ref{fig:1}) adds two key designs upon the standard window-based attention block: (1) adopt a parallel design to combine local-window self-attention and depth-wise convolution, (2) introduce bi-directional interactions across branches. They are proposed to address the limited receptive fields and weak modeling ability issues in local-window self-attention. We first present these two designs then integrate them to build the Mixing Block. Details are described next.

\paragraph{The Parallel Design.} Although performing self-attention inside non-overlapped windows brings computational efficiency\footnote{It has linear computational complexity concerning image size, as shown in Table~\ref{tab:2}.}, it results in a limited receptive field due to no cross-window connections being extracted. Several methods resort to shift~\cite{liu2021swin}, expand~\cite{vaswani2021scaling,yang2021focal}, shuffle~\cite{huang2021shuffle}, or convolution~\cite{huang2021shuffle,YuanFHLZCW21} to model connections across windows. Considering that convolution layers are designed to model local relations, we choose the efficient alternative (depth-wise convolution) as a promising way to connect windows.

Attention then moves to adopt a proper way to combine local-window self-attention and depth-wise convolution. Previous methods~\cite{liu2021swin,vaswani2021scaling,yang2021focal,huang2021shuffle,YuanFHLZCW21} fill the goal by stacking these two operators successively. However, capturing intra-window and cross-window relations in successive steps make these two types of relations less interweaved.

In this paper, we propose a parallel design that enlarges the receptive fields by simultaneously modeling intra-window and cross-window relations. As illustrated in Figure~\ref{fig:1}, local-window self-attention and depth-wise convolution lie in two parallel paths. In detail, they use different window sizes. A $7\times7$ window is adopted in local-window self-attention, following previous works~\cite{vaswani2021scaling,hu2019local,zhao2020exploring,liu2021swin}. While in depth-wise convolution, a smaller kernel size $3\times3$ is applied considering the efficiency\footnote{The results in Table~\ref{tab:9} show that $3\times3$ is a good choice to achieve balance in accuracy and efficiency.}. Moreover, as their FLOPs are different, we adjust the number of channels according to the FLOPs proportion in Table~\ref{tab:2}. Then, their outputs are normalized by different normalization layers~\cite{ioffe2015batch,ba2016layer} and merged by concatenation. The merged feature is sent to the successive Feed-Forward Network (FFN) to mix the learned relations across channels, generating the final output feature.

The parallel design benefits two-folds: First, combining local-window self-attention with depth-wise convolution across branches models connections across windows, addressing the limited receptive fields issue. Second, parallel design models intra-window and cross-window relations simultaneously, providing opportunities for feature interweaving across branches and achieving better feature representation learning.
%##################################################################################################
\begin{figure*}
\centering
\includegraphics[width=1.0\textwidth]{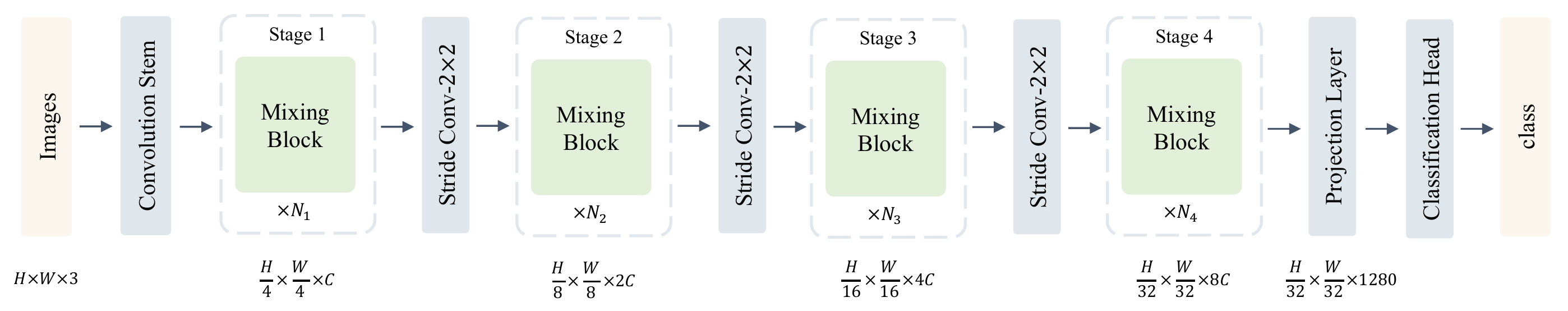}
\caption{\textbf{Overall Architecture of MixFormer.} There are four parts in MixFormer: Convolution Stem, Stages, Projection Layer, and Classification Head. In Convolution Stem, we apply three successive convolutions to increase the channel from $3$ to $C$. In Stages, we stack our Mixing Block in each stage and use stride convolution ($stride=2$) to downsample the feature map. For Projection Layer, we use a linear layer with activation to increase the channels to $1280$. The Classification Head is for the classification task.}
\label{fig:4}\vspace{-3mm}
\end{figure*}
%##################################################################################################

\paragraph{Bi-directional Interactions.} In general, sharing weights limits the modeling ability in the shared dimension. A common way to solve the dilemma is to generate data-dependent weights as done in dynamic networks~\cite{hu2018squeeze,woo2018cbam,chen2021dynamic,li2021involution}. Local-window self-attention computes weights on the fly on the spatial dimension while sharing weights across channels, resulting in the weak modeling ability issue on the channel dimension. We focus on this issue in this subsection.

To enhance the modeling capacity of local-window self-attention on the channel dimension, we try to generate channel-wise dynamic weights~\cite{hu2018squeeze}. Given that depth-wise convolution shares weights on the spatial dimension while focusing on the channel. It can provide complementary clues for local-window self-attention and vice versa. Thus, we propose {\em bi-directional interactions} (in Figure~\ref{fig:1} and Figure~\ref{fig:2}) to enhance modeling ability in the channel and spatial dimension for local-window self-attention and depth-wise convolution respectively. The bi-directional interactions consist of {\em the channel and spatial interaction} among the parallel branches. The information in the depth-wise convolution branch flows to the other branch through the channel interaction, which strengthens the modeling ability in the channel dimension. Meanwhile, the spatial interaction enables spatial relations to flow from the local-window self-attention branch to the other. As a result, the proposed bi-directional interactions provide complementary clues for each other. Next, we present the designs of the channel and spatial interactions in detail.

{\em For the channel interaction}, we follow the design of the SE layer~\cite{hu2018squeeze}, as shown in Figure~\ref{fig:2}. The channel interaction contains one global average pooling layer, followed by two successive $1\times1$ convolution layers with normalization (BN~\cite{ioffe2015batch}) and activation (GELU~\cite{hendrycks2016gaussian}) between them. At last, we use sigmoid to generate attention in the channel dimension. Although our channel interaction shares the same design with the SE layer~\cite{hu2018squeeze}, they differ in two aspects: (1) The input of the attention module is different. The input of our channel interaction comes from another parallel branch, while the SE layer is performed in the same branch. (2) We only apply the channel interaction to the value in the local-window self-attention instead of applying it to the module's output as the SE layer does.

{\em For the spatial interaction}, we also adopt a simple design, which consists of two $1\times1$ convolution layers with followed BN~\cite{ioffe2015batch} and GELU~\cite{hendrycks2016gaussian}. The detailed design is presented in Figure~\ref{fig:2}. These two layers reduce the number of channels to one. At last, a sigmoid layer is used to generate the spatial attention map. Same as we did in the channel interaction, the spatial attention is generated by another branch, where the local-window self-attention module is applied. It has a larger kernel size ($7\times7$) than the depth-wise $3\times3$ convolution and focuses on the spatial dimension, which provides strong spatial clues for the depth-wise convolution branch.

\paragraph{The Mixing Block.} Thanks to the above two designs, we mitigate two core issues in local-window self-attention. We integrate them to build a new transformer block, Mixing Block, upon the standard window attention block. As shown in Figure~\ref{fig:1}, the Mixing Block consists of two efficient operations in a parallel design, bi-directional interactions (Figure~\ref{fig:2}), and an FFN (Feed-Forward Networks)~\cite{vaswani2017attention}. It can be formulated as follow:
\begin{align}
    \hat{X}^{l+1} &= \mathrm{MIX}(\mathrm{LN}(X^{l}), \mathrm{W\text{-}MSA}, \mathrm{CONV}) + X^{l},\\
    X^{l+1} &= \mathrm{FFN}(\mathrm{LN}(\hat{X}^{l+1})) + \hat{X}^{l+1}
\end{align}
Where MIX represents a function that achieves feature mixing between the W-MSA (Window-based Multi-Head Self-Attention) branch and the CONV (Depth-wise Convolution) branch. The MIX function first projects the input feature to parallel branches by two linear projection layers and two norm layers. Then it mixes the features by following the steps shown in Figure~\ref{fig:1} and Figure~\ref{fig:2}. For FFN, we keep it simple and follow previous works~\cite{touvron2021training,liu2021swin}, which is an MLP that consists of two linear layers with one GELU~\cite{hendrycks2016gaussian} between them. Moreover, we also try to add depth-wise convolution as done in PVTv2~\cite{wang2021pvtv2} and HRFormer~\cite{YuanFHLZCW21}, which does not give many improvements over the MLP design (Table~\ref{tab:10}). Thus, to keep the block simple, we use MLP in FFN.

\subsection{MixFormer}
\paragraph{Overall Architecture.} Based on the obtained block, we design an efficient and general-purpose vision transformer, MixFormer, with pyramid feature maps. There are four stages with downsampling rates of $\{4,8,16,32\}$ respectively. MixFormer is a hybrid vision transformer, which uses convolution layers in both stem layers and downsampling layers. Besides, we introduce a projection layer in the tail of the stages. The projection layer increases the feature's channels to $1280$ with a linear layer followed by an activation layer, aiming to preserve more details in the channel before the classification head. It gives a higher performance in classification, especially works well with smaller models. Same design can be found in previous efficient networks, such as MobileNets~\cite{howard2017mobilenets,sandler2018mobilenetv2} and EfficeintNets~\cite{tan2019efficientnet}. The sketch of our MixFormer is given in Figure~\ref{fig:4}.

\paragraph{Architecture Variants.} We stack the blocks in each stage manually and format several models in different sizes, whose computational complexities ranges from $0.7$G (B1) to $3.6$G (B4). The number of blocks in different stages is set by following a recipe: putting more blocks in the last two stages, which is roughly verified in Table~\ref{tab:11}. As shown in Table~\ref{tab:3}, we present the detailed settings of the models. 
%##################################################################################################
\begin{table}
\tablestyle{1.pt}{1.1}
\begin{tabular}{x{65}x{35}x{50}x{55}}
\shline
Models & \#Channels   & \#Blocks & \#Heads\\
\hline
MixFormer-B1   &  $C=32$ & $[1, 2, 6, 6]$ & $[2, 4, 8, 16]$\\
MixFormer-B2   &  $C=32$ & $[2, 2, 8, 8]$ & $[2, 4, 8, 16]$\\
MixFormer-B3   &  $C=48$ & $[2, 2, 8, 6]$ & $[3, 6, 12, 24]$\\
MixFormer-B4   &  $C=64$ & $[2, 2, 8, 8]$ & $[4, 8, 16, 32]$\\
% MixFormer-B5   &  $C=96$ & $[1, 2, 8, 6]$ & $[6, 12, 24, 48]$\\
\shline
\end{tabular}
\caption{\textbf{Architecture Variants.} Detailed configurations of architecture variants of MixFormer.}
\label{tab:3}\vspace{-4mm}
\end{table}
%##################################################################################################
%-------------------------------------------------------------------------
\section{Experiments}
\label{sec:exp}
%-------------------------------------------------------------------------
We validate MixFormer on ImageNet-1K~\cite{deng2009imagenet}, MS COCO~\cite{lin2014microsoft}, and ADE20k~\cite{zhou2019semantic}. We first present the accuracy on image classification. Then we do transfer learning to evaluate the models on three main tasks: object detection, instance segmentation, and semantic segmentation. Besides, ablations of different design modules in MixFormer and results with more vision tasks are provided.

\subsection{Image Classification}
\paragraph{Setup.} We first verify our method by classification on ImageNet-1K~\cite{deng2009imagenet}. To make a fair comparison with previous works~\cite{touvron2021training,wang2021pyramid,liu2021swin}, we train all models for $300$ epochs with an image size of $224\times224$ and report Top-1 validation accuracy. We apply an AdamW optimizer using a cosine decay schedule. By following the rule that smaller models need less regularization, we adjust the training settings gently when training models in different sizes. Details are in Appendix~\ref{Appendix:C}.
%##################################################################################################
\begin{table}[t]
\centering
\tablestyle{5.pt}{1.1}
\begin{tabular}{cccc}
\shline
\multicolumn{1}{c|}{Method} &  \multicolumn{1}{l|}{\#Params} & \multicolumn{1}{l|}{FLOPs} & \multicolumn{1}{l}{Top-1} \\ 
\midrule
\multicolumn{4}{c}{ConvNets} \\  \midrule
\multicolumn{1}{c|}{RegNetY-0.8G~\cite{radosavovic2020designing}} & \multicolumn{1}{c|}{6M} & \multicolumn{1}{c|}{0.8G} & \multicolumn{1}{c}{76.3}\\
\multicolumn{1}{c|}{RegNetY-1.6G~\cite{radosavovic2020designing}}
& \multicolumn{1}{c|}{11M} & \multicolumn{1}{c|}{1.6G}  & \multicolumn{1}{c}{78.0}\\
\multicolumn{1}{c|}{RegNetY-4G~\cite{radosavovic2020designing}}
& \multicolumn{1}{c|}{21M} & \multicolumn{1}{c|}{4.0G} & \multicolumn{1}{c}{80.0}\\ 
\multicolumn{1}{c|}{RegNetY-8G~\cite{radosavovic2020designing}}
& \multicolumn{1}{c|}{39M} & \multicolumn{1}{c|}{8.0G}  & \multicolumn{1}{c}{81.7}\\
\multicolumn{1}{c|}{EffNet-B1~\cite{tan2019efficientnet}}
& \multicolumn{1}{c|}{8M} & \multicolumn{1}{c|}{0.7G} & \multicolumn{1}{c}{79.1}\\
\multicolumn{1}{c|}{EffNet-B2~\cite{tan2019efficientnet}}
& \multicolumn{1}{c|}{9M} & \multicolumn{1}{c|}{1.0G} & \multicolumn{1}{c}{80.1}\\
\multicolumn{1}{c|}{EffNet-B3~\cite{tan2019efficientnet}}
& \multicolumn{1}{c|}{12M} & \multicolumn{1}{c|}{1.8G} & \multicolumn{1}{c}{81.6}\\
\multicolumn{1}{c|}{EffNet-B4~\cite{tan2019efficientnet}}
& \multicolumn{1}{c|}{19M} & \multicolumn{1}{c|}{4.2G} & \multicolumn{1}{c}{82.9}\\
\hline
\multicolumn{4}{c}{Vision Transformers}\\
\hline
\multicolumn{1}{c|}{DeiT-T~\cite{touvron2021training}}& \multicolumn{1}{c|}{6M}
& \multicolumn{1}{c|}{1.3G} & \multicolumn{1}{c}{72.2} \\
\multicolumn{1}{c|}{DeiT-S~\cite{touvron2021training}}
&\multicolumn{1}{c|}{22M} & \multicolumn{1}{c|}{4.6G}
& \multicolumn{1}{c}{79.9}\\
\multicolumn{1}{c|}{DeiT-B~\cite{touvron2021training}}
&\multicolumn{1}{c|}{87M} & \multicolumn{1}{c|}{17.5G}
& \multicolumn{1}{c}{81.8}\\
\multicolumn{1}{c|}{PVT-T~\cite{wang2021pyramid}} & \multicolumn{1}{c|}{13M}  
& \multicolumn{1}{c|}{1.8G} &\multicolumn{1}{c}{75.1} \\
\multicolumn{1}{c|}{PVT-S~\cite{wang2021pyramid}}
& \multicolumn{1}{c|}{25M} & \multicolumn{1}{c|}{3.8G}
& \multicolumn{1}{c}{79.8}\\
\multicolumn{1}{c|}{PVT-M~\cite{wang2021pyramid}}& \multicolumn{1}{c|}{44M} & \multicolumn{1}{c|}{6.7G}
& \multicolumn{1}{c}{81.2}\\
\multicolumn{1}{c|}{PVT-L~\cite{wang2021pyramid}}& \multicolumn{1}{c|}{61M}
& \multicolumn{1}{c|}{9.8G} & \multicolumn{1}{c}{81.7}\\
\multicolumn{1}{c|}{CvT-13\cite{wu2021cvt}}& \multicolumn{1}{c|}{20M}
& \multicolumn{1}{c|}{4.5G} & \multicolumn{1}{c}{81.6}\\
\multicolumn{1}{c|}{CvT-21\cite{wu2021cvt}} & \multicolumn{1}{c|}{32M}
& \multicolumn{1}{c|}{7.1G} & \multicolumn{1}{c}{82.5}\\
\multicolumn{1}{c|}{TwinsP-S~\cite{chu2021twins}}  & \multicolumn{1}{c|}{24M}
& \multicolumn{1}{c|}{3.8G} & \multicolumn{1}{c}{81.2}\\
\multicolumn{1}{c|}{DS-Net-S\cite{mao2021dual}} & \multicolumn{1}{c|}{23M}
& \multicolumn{1}{c|}{3.5G} & \multicolumn{1}{c}{82.3}\\
\hline
\multicolumn{1}{c|}{Swin-T~\cite{liu2021swin}}& \multicolumn{1}{c|}{29M}
& \multicolumn{1}{c|}{4.5G} & \multicolumn{1}{c}{81.3}\\
\multicolumn{1}{c|}{Swin-S~\cite{liu2021swin}}& \multicolumn{1}{c|}{50M}
& \multicolumn{1}{c|}{8.7G} & \multicolumn{1}{c}{83.0}\\
\multicolumn{1}{c|}{Twins-S~\cite{chu2021twins}}& \multicolumn{1}{c|}{24M}
& \multicolumn{1}{c|}{2.9G} & \multicolumn{1}{c}{81.7}\\
\multicolumn{1}{c|}{LG-T\cite{li2021local}}& \multicolumn{1}{c|}{33M}
& \multicolumn{1}{c|}{4.8G} & \multicolumn{1}{c}{82.1}\\
\multicolumn{1}{c|}{Focal-T\cite{yang2021focal}}& \multicolumn{1}{c|}{29M}       & \multicolumn{1}{c|}{4.9G} & \multicolumn{1}{c}{82.2}\\
\multicolumn{1}{c|}{Shuffle-T\cite{huang2021shuffle}}& \multicolumn{1}{c|}{29M}       & \multicolumn{1}{c|}{4.6G} & \multicolumn{1}{c}{82.5}\\
\hline
\multicolumn{1}{c|}{MixFormer-B1 (\textbf{Ours})}  & \multicolumn{1}{c|}{8M}  & \multicolumn{1}{c|}{0.7G}  &\multicolumn{1}{c}{78.9}\\
\multicolumn{1}{c|}{MixFormer-B2 (\textbf{Ours})}
& \multicolumn{1}{c|}{10M}  & \multicolumn{1}{c|}{0.9G}
& \multicolumn{1}{c}{80.0}\\
\multicolumn{1}{c|}{MixFormer-B3 (\textbf{Ours})}  & \multicolumn{1}{c|}{17M} & \multicolumn{1}{c|}{1.9G} 
& \multicolumn{1}{c}{81.7}\\
\multicolumn{1}{c|}{MixFormer-B4 (\textbf{Ours})}  & \multicolumn{1}{c|}{35M}  & \multicolumn{1}{c|}{3.6G}  & \multicolumn{1}{c}{83.0}\\
\shline
\end{tabular}
\caption{\textbf{Classification accuracy on the ImageNet validation set.} Performances are measured with a single $224\times224$ crop. ``Params'' refers to the number of parameters. ``FLOPs'' is calculated under the input scale of $224\times 224$.}
\label{tab:4}\vspace{-5mm}
\end{table}
%##################################################################################################
%##################################################################################################
\begin{table*}
\centering
% \scriptsize
% \footnotesize
\begin{center}
\resizebox{0.9\linewidth}{!}{
\setlength{\tabcolsep}{1.5mm}{
\begin{tabular}{c@{\hspace{3.2pt}}|c@{\hspace{3.2pt}}|c|c|c|c|c|c|c|c|c|c|c|c|c}
\shline
\multirow{2}{*}{Backbones} & \multirow{2}{*}{\#Params} & \multirow{2}{*}{FLOPs} & \multicolumn{6}{c|}{Mask R-CNN 1x schedule} & \multicolumn{6}{c}{Mask R-CNN 3x + MS schedule}\\
\cmidrule{4-15}
& & & $AP^b$ & $AP^b_{50}$ & $AP^b_{75}$ & $AP^m$ & $AP^m_{50}$ & $AP^m_{75}$ & $AP^b$ & $AP^b_{50}$ & $AP^b_{75}$ & $AP^m$ & $AP^m_{50}$ & $AP^m_{75}$ \\
\hline
ResNet18~\cite{he2016deep} & 31M & - & 34.0 & 54.0 & 36.7 & 31.2 & 51.0 & 32.7 & 36.9 & 57.1 & 40.0 & 33.6 & 53.9 & 35.7\\
ResNet50~\cite{he2016deep} & 44M & 260G  & 38.0 & 58.6 & 41.4 & 34.4 & 55.1 & 36.7
& 41.0 & 61.7 & 44.9 & 37.1 & 58.4 & 40.1 \\
ResNet101~\cite{he2016deep} & 63M & 336G  & 40.4 & 61.1 & 44.2 & 36.4 & 57.7 & 38.8
& 42.8 & 63.2 & 47.1 & 38.5 & 60.1 & 41.3 \\
ResNeXt101-64$\times$4d~\cite{xie2017aggregated} & 101M & 493G  & 42.8 & 63.8 & 47.3 & 38.4 & 60.6 & 41.3
& 44.4 & 64.9 & 48.8 & 39.7 & 61.9 & 42.6 \\
\hline
PVT-T~\cite{wang2021pyramid} & 33M & - & 36.7 & 59.2 & 39.3 & 35.1 & 56.7 & 37.3 & 39.8 & 62.2 & 43.0 & 37.4 & 59.3 & 39.9\\
PVT-S~\cite{wang2021pyramid} & 44M & 245G  & 40.4 & 62.9 & 43.8 & 37.8 & 60.1 & 40.3
& 43.0 & 65.3 & 46.9 & 39.9 & 62.5 & 42.8 \\
PVT-M~\cite{wang2021pyramid} & 64M & 302G  & 42.0 & 64.4 & 45.6 & 39.0 & 61.6 & 42.1
& 44.2 & 66.0 & 48.2 & 40.5 & 63.1 & 43.5 \\
PVT-L~\cite{wang2021pyramid} & 81M & 364G  & 42.9 & 65.0 & 46.6 & 39.5 & 61.9 & 42.5
& 44.5 & 66.0 & 48.3 & 40.7 & 63.4 & 43.7 \\
TwinsP-S~\cite{chu2021twins}           & 44M & 245G  & 42.9 & 65.8 & 47.1 & 40.0 & 62.7 & 42.9
& 46.8 & 69.3 & 51.8 & 42.6 & 66.3 & 46.0  \\
DS-Net-S~\cite{mao2021dual}           & 43M & -  & 44.3 & - & - & 40.2 & - & -
& - & - & - & - & - & -  \\
Swin-T~\cite{liu2021swin}              & 48M & 264G  & 42.2 & 64.6 & 46.2 & 39.1 & 61.6 & 42.0
& 46.0 & 68.2 & 50.2 & 41.6 & 65.1 & 44.8 \\
Twins-S~\cite{chu2021twins}            & 44M & 228G  & 43.4 & 66.0 & 47.3 & 40.3 & 63.2 & 43.4
& 46.8 & 69.2 & 51.2 & 42.6 & 66.3 & 45.8 \\
Focal-T~\cite{yang2021focal}            & 49M & 291G  & - & - & - & - & - & -
& 47.2 & 69.4 & 51.9 & 42.7 & 66.5 & 45.9 \\
Shuffle-T~\cite{huang2021shuffle}    & 48M & 268G  & - & - & - & - & - & - &46.8&68.9&51.5&42.3&66.0&45.6\\
\hline
MixFormer-B1(\textbf{Ours}) & 26M & 183G  & 40.6 & 62.6 & 44.1 & 37.5 & 59.7 & 40.0
& 43.9 & 65.6 & 48.1 & 40.0 & 62.9 & 42.9 \\
MixFormer-B2(\textbf{Ours}) & 28M & 187G  & 41.5 & 63.3 & 45.2 & 38.3 & 60.6 & 41.2
& 45.1 & 66.9 & 49.2 & 40.8 & 64.1 & 43.6 \\
MixFormer-B3(\textbf{Ours}) & 35M & 207G  & 42.8 & 64.5 & 46.7 & 39.3 & 61.8 & 42.2
& 46.2 & 68.1 & 50.5 & 41.9 & 65.6 & 45.0 \\
MixFormer-B4(\textbf{Ours})  & 53M & 243G  & \textbf{45.1} & \textbf{67.1} & \textbf{49.2} & \textbf{41.2} & \textbf{64.3} & \textbf{44.1}
& \textbf{47.6} & \textbf{69.5} & \textbf{52.2} & \textbf{43.0} & \textbf{66.7} & \textbf{46.4}\\
\shline
\end{tabular}
}}
\end{center}\vspace{-2mm}
\caption{\textbf{COCO detection and segmentation with the Mask R-CNN.} The performances are reported on the COCO {\em val} split under $1\times$ and $3\times$ schedules. The FLOPs (G) are measured at resolution $800\times 1280$, and all models are pre-trained on the ImageNet-1K~\cite{deng2009imagenet}. In the table, '-' means that the result is not reported by the original paper.}
\label{tab:5}\vspace{-2mm}
\end{table*}
%##################################################################################################

\paragraph{Results.} Table~\ref{tab:4} compares our MixFormer with efficient ConvNets~\cite{tan2019efficientnet,radosavovic2020designing} and various Vision Transformers~\cite{touvron2021training,wang2021pyramid,wu2021cvt,liu2021swin,li2021local,yang2021focal, huang2021shuffle}. MixFormer performs on par with EfficientNet~\cite{tan2019efficientnet} and outperforms RegNet~\cite{radosavovic2020designing} by significant margins under various computational budgets (from B1 to B4). We note that it is {\em nontrivial} to achieve such results for vision transformer-based models, especially on small models (FLOPs $< 1.0$G ). Previous works such as DeiT~\cite{touvron2021training} and PVT~\cite{wang2021pyramid} show dramatic performance drops when reducing model complexities ($-7.7\%$ from DeiT-S to DeiT-T and $-4.7\%$ from PVT-S to PVT-T). Compared with Swin Transformer~\cite{liu2021swin} and its variants~\cite{li2021local,yang2021focal, huang2021shuffle}, MixFormer shows better performance with less computational costs. In detail, MixFormer-B4 achieves $83.0\%$ Top-1 accuracy with only $3.6$G FLOPs. It outperforms Swin-T~\cite{liu2021swin} by $1.7\%$ while saving $20\%$ computational costs and gives comparable results with Swin-S~\cite{liu2021swin} but being $2.4\times$ efficient. The competitive advantage of MixFormer maintains when it comes to LG-Transformer~\cite{li2021local}, Focal Transformer~\cite{yang2021focal} and Shuffle Transformer~\cite{huang2021shuffle}.
Moreover, our MixFormer also scales well to smaller and larger models. More results are provided in Appendix~\ref{Appendix:A}.
%##################################################################################################
\begin{table}
\begin{center}
% \scriptsize
% \footnotesize
\resizebox{\linewidth}{!}{
\setlength{\tabcolsep}{1.5mm}{
\addtolength{\tabcolsep}{-2.5pt}
\begin{tabular}{c|c|c|c|c|c|c|c|c}
\shline
Backbones & \#Params & FLOPs & $AP^b$ & $AP^b_{50}$ & $AP^b_{75}$ & $AP^m$ & $AP^m_{50}$ & $AP^m_{75}$ \\
\hline
ResNet50~\cite{he2016deep} & 82M & 739G & 46.3 & 64.3 & 50.5 & 40.1 & 61.7 & 43.4\\
Swin-T~\cite{liu2021swin} & 86M & 745G & 50.5 & 69.3 & 54.9 & 43.7 & 66.6 & 47.1\\
Shuffle-T~\cite{huang2021shuffle} & 86M & 746G & 50.8 & 69.6 & 55.1 & 44.1 & 66.9 & 48.0\\
MixFormer-B4(\textbf{Ours})  & 91M & 721G  & \textbf{51.6} & \textbf{70.5} & \textbf{56.1} & \textbf{44.9} & \textbf{67.9} & \textbf{48.7} \\
\shline
\end{tabular}
}}
\end{center}\vspace{-4mm}
\caption{\textbf{COCO detection and segmentation with the Cascade Mask R-CNN.} The performances are reported on the COCO {\em val} split under a $3\times$ schedule. Results show consistent improvements of MixFormer over Swin Transformer.}
\label{tab:6}\vspace{-2mm}
\end{table}
%##################################################################################################

\subsection{Object Detection and Instance Segmentation}
\paragraph{Setup.} We validate the effectiveness of MixFormer on downstream tasks. We train Mask R-CNN~\cite{he2017mask} on the COCO2017 {\em train} split and evaluate the models on the {\em val} split. Two training schedules ($1\times$ and $3\times$) are adopted to show a consistent comparison with previous methods~\cite{he2016deep,touvron2021training,liu2021swin,li2021local,yang2021focal}. For the $1\times$ schedule, we train for $12$ epochs with a single size (resizing the shorter side to $800$ while keeping its longer side no more than $1333$)~\cite{he2017mask}. While in $3\times$ schedule (36 epochs), we use multi-scale training by randomly resizing the shorter side to the range of $[480, 800]$ (See Appendix~\ref{Appendix:C} for more details). Expect for Mask R-CNN~\cite{he2017mask}, we also provide comparisons with previous works based on a stronger model, Cascade Mask R-CNN~\cite{cai2018cascade,he2017mask}, where a $3\times$ schedule is conducted.
%##################################################################################################
\begin{table}
\begin{center}
% \scriptsize
% \footnotesize
\resizebox{\linewidth}{!}{
\setlength{\tabcolsep}{1.5mm}{
\addtolength{\tabcolsep}{-2.5pt}
\begin{tabular}{cc|cc|cc}
\shline
Backbone & Method &  \#Params & FLOPs & mIoU$_{ss}$ & mIoU$_{ms}$\\
\hline
ResNet-101~\cite{he2016deep} & DANet~\cite{fu2019dual} &  69M & 1119G & 43.6 &45.2  \\
ResNet-101~\cite{he2016deep} & DLab.v3+~\cite{chen2018deeplabv3+} &  63M & 1021G  & 45.1 & 46.7  \\
ResNet-101~\cite{he2016deep} & ACNet~\cite{fu2019adaptive} &  - & - & 45.9 & -  \\
ResNet-101~\cite{he2016deep} & DNL~\cite{yin2020disentangled} &  69M & 1249G & 46.0 & -  \\
ResNet-101~\cite{he2016deep} & OCRNet~\cite{yuan2020object} &  56M & 923G & - & 45.3 \\
ResNet-101~\cite{he2016deep} & UperNet~\cite{xiao2018unified} &  86M & 1029G  & 43.8 & 44.9 \\
HRNet-w48~\cite{wang2020deep} & OCRNet~\cite{yuan2020object} &  71M & 664G & -  & 45.7  \\
\hline
DeiT-S~\cite{touvron2021training}$^\dag$ & UperNet~\cite{xiao2018unified} &  52M & 1099G  & 44.0 & - \\
TwinsP-S~\cite{chu2021twins}&UperNet~\cite{xiao2018unified}           & 55M & 919G  & 46.2 & 47.5 \\
Swin-T~\cite{liu2021swin}  &UperNet~\cite{xiao2018unified}  &  60M & 945G & 44.5 & 45.8 \\
Twins-S~\cite{chu2021twins}        &UperNet~\cite{xiao2018unified}    & 54M & 901G  & 46.2 & 47.1\\
LG-T~\cite{li2021local} & UperNet~\cite{xiao2018unified} &  64M & 957G &- &  45.3  \\
Focal-T~\cite{yang2021focal} & UperNet~\cite{xiao2018unified} &  62M & 998G & 45.8 & 47.0 \\
Shuffle-T~\cite{huang2021shuffle} & UperNet~\cite{xiao2018unified} &  60M & 949G & 46.6 & 47.6 \\
\hline
MixFormer-B1(\textbf{Ours})  & UperNet~\cite{xiao2018unified} &  35M & 854G & 42.0 & 43.5\\
MixFormer-B2(\textbf{Ours})  & UperNet~\cite{xiao2018unified} &  37M & 859G & 43.1 & 43.9\\
MixFormer-B3(\textbf{Ours})  & UperNet~\cite{xiao2018unified} &  44M & 880G & 44.5 & 45.5\\
MixFormer-B4(\textbf{Ours})  & UperNet~\cite{xiao2018unified} &  63M & 918G & \textbf{46.8} & \textbf{48.0}\\
\shline
\end{tabular}

}}
\end{center}\vspace{-2mm}
\caption{\textbf{ADE20K semantic segmentation.} We report mIoU on the ADE20K~\cite{zhou2019semantic} {val} split with single scale (ss) testing and multi-scale (ms) testing . A resolution $512\times 2048$ is used to measure the FLOPs (G) in various models.}
\label{tab:7}\vspace{-6mm}
\end{table}
%##################################################################################################

\paragraph{Comparison on Mask R-CNN.} Table~\ref{tab:5} shows that MixFormer consistently outperforms other competitors~\cite{he2016deep,wang2021pyramid,wu2021cvt,liu2021swin,li2021local,yang2021focal} under various model sizes with Mask R-CNN~\cite{he2017mask}. In particular, MixFormer-B4 achieves \text{+}$\pmb{2.9}$(\text{+}$\pmb{1.6}$) higher box mAP and \text{+}$\pmb{2.1}$(\text{+}$\pmb{1.4}$) higher mask mAP than the Swin-T~\cite{liu2021swin} baseline with $1\times$ ($3\times$) schedule.
Moreover, MixFormer keeps its efficiency in detection and instance segmentation, enabling higher performance with less computational costs than other networks~\cite{he2016deep,liu2021swin}. 
It is a surprise that our MixFormer-B1 (only with 0.7G) performs strongly with Mask R-CNN ($1\times$), which exceeds ResNet-50 (with 4.1G)~\cite{he2016deep} by 2.3 box mAP and 2.9 mask mAP. The results suggest that implications for designing high-performance small models on detection are highlighted in MixFormer.

\paragraph{Comparison on Cascade Mask R-CNN.} We also evaluate MixFormer with Cascade Mask R-CNN~\cite{he2017mask,cai2018cascade}, which is a stronger variant of Mask R-CNN~\cite{he2017mask}. MixFormer-B4 provides robust improvements over Swin-T~\cite{liu2021swin} (Table~\ref{tab:6}) regardless of different detectors, as it shows similar gains (+1.1/1.2 box/mask mAP \text{v.s.} +1.6/1.4 box/mask mAP) with the ones on Mask R-CNN ($3\times$) (Table~\ref{tab:5}).

\subsection{Semantic Segmentation}
\paragraph{Setup.} Our experiments are conducted on ADE20K~\cite{zhou2019semantic} using UperNet~\cite{xiao2018unified}. For training recipes, we mainly follow the settings in~\cite{liu2021swin}. We report mIoU of our models in single scale testing (ss) and multi-scale testing (ms). Details are provided in Appendix~\ref{Appendix:C}.

\paragraph{Results.} In Table~\ref{tab:7}, MixFormer-B4 consistently achieves better mIoU performance than previous networks.
It seems that the connections across windows and dimensions in the Mixing Block provide more benefits on semantic segmentation as the gains are larger than the ones on detection tasks (Table~\ref{tab:5},Table~\ref{tab:6}) with the same backbones. In particular, MixFormer-B4 outperforms Swin-T~\cite{liu2021swin} by $\pmb{2.2}$ mIoU.

Moreover, other variants of MixFormer (from B1 to B3) also achieve higher performance while being more efficient than previous networks. Notably, MixFormer-B3 obtains $45.5$ mIoU (comparable with Swin-T~\cite{liu2021swin} but less FLOPs), which achieves on par results with OCRNet~\cite{yuan2020object} with HRNet-W48~\cite{wang2020deep} ($45.7$ mIoU). Note that HRNet~\cite{wang2020deep} is carefully designed to aggregate the features in different stages, while MixFormer simply constructs pyramid feature maps, indicating the strong potential for further improvements on dense prediction tasks.

\subsection{Ablation Study}
%##################################################################################################
\begin{table}
\small
\centering
\addtolength{\tabcolsep}{-4.5pt}
\begin{tabular}{c|cc|cc|cc|c}
\shline
\multirow{2}{*}{Parallel} & \multicolumn{2}{c|}{Interactions} & \multicolumn{2}{c|}{ImageNet} & \multicolumn{2}{c|}{COCO} & \multicolumn{1}{c}{ADE20k} \\
& Channel & Spatial & Top-1 & Top-5  & AP$^\text{box}$ & AP$^\text{mask}$ & mIoU \\
\hline
& & & 77.4 & 93.8 & 38.2 & 35.7 & 38.9 \\
\checkmark & & & 78.1 & 94.1 & 39.4 & 36.6 & 39.8 \\
\checkmark & \checkmark &  & 78.3 & 94.1 & 40.1 & 37.1 & 40.6\\
\checkmark & & \checkmark & 78.3 & 94.1 & 39.7 & 36.6 & 40.5\\
\checkmark & \checkmark & \checkmark & \textbf{78.4} & \textbf{94.3} & \textbf{40.3} & \textbf{37.3} & \textbf{40.9} \\
\hline
\multicolumn{3}{c|}{$\Delta$} & \cgaphl{+}{1.0} & \cgaphl{+}{0.5} & \cgaphl{+}{2.1} & \cgaphl{+}{1.6} & \cgaphl{+}{2.0} \\
\shline
\end{tabular}
\caption{\textbf{Parallel Design with Bi-directional Interactions.} The baseline model in this table adopts a successive design and has no interactions in the block.}
\label{tab:8}\vspace{-2mm}
\end{table}
%##################################################################################################
%##################################################################################################
\begin{table}
\small
\centering
\addtolength{\tabcolsep}{-2.5pt}
\begin{tabular}{c|cc|cc|c}
\shline
\multirow{2}{*}{Window Sizes} & \multicolumn{2}{c|}{ImageNet} & \multicolumn{2}{c|}{COCO} & \multicolumn{1}{c}{ADE20k} \\
 & Top-1 & Top-5  & AP$^\text{box}$ & AP$^\text{mask}$ & mIoU \\
\hline
$1\times1$ & 77.1 & 93.6 & 36.3 & 34.3  & 37.6 \\
\pmb{$3\times3$} & \textbf{78.4} & \textbf{94.3} & \textbf{40.3} & \textbf{37.3}  & \textbf{40.9} \\
$5\times5$ & 78.4 & 94.3 & 40.3 & 37.2 & 40.8\\
% $7\times7$ & 78.60 & 94.22 & 40.2 & 37.0 & 41.01 \\
\shline
\end{tabular}
\caption{\textbf{Window Sizes in DwConv.} We investigate various window sizes for DwConv. MixFormer uses the $3\times3$ window size for DwConv by default.}
\label{tab:9}\vspace{-2mm}
\end{table}
%##################################################################################################
\paragraph{Setup.} We provide ablations with respect to our designs on MixFormer-B1. We report all variations of different designs on ImageNet-1K~\cite{deng2009imagenet} classification, COCO~\cite{lin2014microsoft} detection and segmentation, and ADE20K~\cite{zhou2019semantic} semantic segmentation. To make quick evaluations, we only train MixFormer-B1 for $200$ epochs on ImageNet-1K~\cite{deng2009imagenet}. Then, the pre-trained models are adopted by Mask R-CNN~\cite{he2017mask} ($1\times$) on MS COCO~\cite{lin2014microsoft} and UperNet~\cite{xiao2018unified} ($160k$) on ADE20K~\cite{zhou2019semantic}. Note that, the differences in pre-train models provide slightly different results with the ones in Table~\ref{tab:5} and Table~\ref{tab:7}.

\paragraph{Ablation: Parallel or Not.} Table~\ref{tab:8} provides the comparison of the ways (successive design or parallel design) to combine local-window self-attention and depth-wise convolution. Our parallel design consistently outperforms the successive design across various vision tasks, which verifies the hypothesis that the parallel design enables better feature representation learning in Section~\ref{sec:intro}. The models below use parallel design by default.
%##################################################################################################
\begin{table}
\small
\centering
\addtolength{\tabcolsep}{-3.5pt}
\begin{tabular}{c|cc|cc|c}
\shline
\multirow{2}{*}{Techniques} & \multicolumn{2}{c|}{ImageNet} & \multicolumn{2}{c|}{COCO} & \multicolumn{1}{c}{ADE20k} \\
 & Top-1 & Top-5  & AP$^\text{box}$ & AP$^\text{mask}$ & mIoU \\
\hline
MixFormer-B1(\textbf{Ours})  & 78.4 & 94.3 & 40.3 & 37.3  & 40.9 \\
+shifted windows & 78.3 & 94.1 & 40.5 & 37.3 & 40.7 \\
+DwConv in FFN & 78.6 & 94.4 & 40.5 & 37.4 & 40.9\\
\shline
\end{tabular}
\caption{\textbf{Other Techniques.} We combine two techniques with our MixFormer. When inserting DwConv in FFN, we only consider $3\times3$ DwConv.}
\label{tab:10}\vspace{-2mm}
\end{table}
%##################################################################################################
%##################################################################################################
\begin{table}
\small
\centering
\addtolength{\tabcolsep}{-5.5pt}
\begin{tabular}{c|c|cc|cc|c}
\shline
\#Blocks & \multirow{2}{*}{FLOPs}& \multicolumn{2}{c|}{ImageNet} & \multicolumn{2}{c|}{COCO} & \multicolumn{1}{c}{ADE20k} \\
\#Channels & & Top-1 & Top-5  & AP$^\text{box}$ & AP$^\text{mask}$ & mIoU \\
\hline
$[2,2,8,2]$  &\multirow{2}{*}{0.9G} & \multirow{2}{*}{77.7} & \multirow{2}{*}{93.9} & \multirow{2}{*}{40.1} & \multirow{2}{*}{37.3}  & \multirow{2}{*}{40.6} \\
$[32,64,160,256]$ & & & & & & \\
\hline
$[2,2,6,4]$ &\multirow{2}{*}{0.9G} & \multirow{2}{*}{77.5} & \multirow{2}{*}{93.7} & \multirow{2}{*}{39.6} & \multirow{2}{*}{36.7}  & \multirow{2}{*}{39.8} \\
$[32,64,128,256]$ & & & & & & \\
\hline
$[1,2,6,2]$ &\multirow{2}{*}{0.8G} & \multirow{2}{*}{77.2}  & \multirow{2}{*}{93.5} & \multirow{2}{*}{39.3} & \multirow{2}{*}{36.6}  & \multirow{2}{*}{40.4} \\
$[32,64,160,320]$ & & & & & & \\
\hline
$[1,2,6,6]$ &\multirow{2}{*}{\pmb{0.7G}} & \multirow{2}{*}{\textbf{78.4}} & \multirow{2}{*}{\textbf{94.3}} & \multirow{2}{*}{\textbf{40.3}} & \multirow{2}{*}{\textbf{37.3}}  & \multirow{2}{*}{\textbf{40.9}} \\
$[32,64,128,256]$ & & & & & & \\
\shline
\end{tabular}
\caption{\textbf{Number of Blocks in Stages.} In the table, the first two models and the last two models share similar computational complexities with each other.}
\label{tab:11}\vspace{-3mm}
\end{table}
%##################################################################################################

\paragraph{Ablation: Bi-directional Interactions.} Table~\ref{tab:8} shows the results of the proposed interactions. According to the results, we see that both channel and spatial interactions outperform the model without interactions across all different vision tasks. Combining two interactions promotes better performance, resulting in consistent improvements by $0.3\%$ Top-1 accuracy on ImageNet-1K, $0.9/0.7$ box/mask mAP on COCO, and $1.1$ mIoU on ADE20K. Given that we only use simple and light-weighted designs for bi-directional interactions, the gains are nontrivial, which indicates the effectiveness of providing complementary clues for local-window self-attention and depth-wise convolution.

\paragraph{Ablation: Window Sizes in DwConv.} Table~\ref{tab:9} shows that the performance will drop significantly on various vision tasks ($-1.3$ Top-1 accuracy on ImageNet-1K, $-4.0/-3.0$ box/mask mAP on COCO, and $-3.3$ mIoU on ADE20K) if we reduce the window size of the depth-wise convolution from $3\times3$ to $1\times1$. This phenomenon means that it's necessary for depth-wise convolution to use a window size  (at least $3\times3$) with the ability to connect across-window. Besides, when we increase the window size to $5\times5$, no clear further gains are observed. Thus, we use a window size of $3\times3$ regarding the efficiency.

\paragraph{Ablation: Other Techniques.} We also investigate other designs in MixFormer, including applying shifted windows and inserting $3\times3$ depth-wise convolution in FFN, which play significant roles in previous works~\cite{liu2021swin,YuanFHLZCW21}. As presented in Table~\ref{tab:10}, shifted window fails to provide gains over MixFormer. We hypothesize that the depth-wise convolution builds connections among windows, removing the need for shift operation. Besides, although inserting $3\times3$ depth-wise convolution in FFN can provide further gains, the room for improvements is limited with MixFormer. Thus, we use MLP in FFN by default.

\paragraph{Ablation: Number of Blocks in Stages.} Previous works usually put more blocks in the third stage and greatly increase the number of blocks in that stage when scaling models~\cite{he2016deep,wang2021pyramid,liu2021swin}. We show an alternative way that can achieve the goal. We roughly conduct experiments on the way of stacking blocks. In Table~\ref{tab:11}, we achieve slightly higher performance on various vision tasks under less computational complexities by putting more blocks in both the last two stages. We follow this recipe to build our MixFormer.
%##################################################################################################
\begin{table}
\small
\centering
\addtolength{\tabcolsep}{-4.5pt}
\begin{tabular}{c|cccccc}
\shline
\multirow{2}{*}{Backbones} & \multicolumn{6}{c}{COCO keypoint detection} \\
 & \multicolumn{2}{l}{AP$^\text{kp}$} & \multicolumn{2}{l}{AP$_{50}^\text{kp}$} & \multicolumn{2}{l}{AP$_{75}^\text{kp}$} \\
\hline
ResNet50~\cite{he2016deep} & 71.8 & ~ & 89.8 & ~ & 79.5 & ~\\
Swin-T~\cite{liu2021swin} & 74.2 & ~ & 92.5 & ~ & 82.5 & ~\\
HRFormer-S~\cite{liu2021swin} &74.5 & ~ & 92.3 & ~ & 82.1 & ~\\
MixFormer-B4(\textbf{Ours}) & \textbf{75.3} & \cgaphlp{+}{1.1} & \textbf{93.5} & \cgaphlp{+}{1.0} & \textbf{83.5} & \cgaphlp{+}{1.0}\\
\shline
\end{tabular}
\\[2mm]
\addtolength{\tabcolsep}{-0pt}
\begin{tabular}{c|cccccc}
\shline
\multirow{2}{*}{Backbones} & \multicolumn{6}{c}{LVIS Instance Segmentation} \\
 & \multicolumn{2}{l}{AP$^\text{mask}$} & \multicolumn{2}{l}{AP$_{50}^\text{mask}$} & \multicolumn{2}{l}{AP$_{75}^\text{mask}$} \\
\hline
ResNet50~\cite{he2016deep} & 21.7 & ~ & 34.3 & ~ & 23.0 & ~\\
Swin-T~\cite{liu2021swin} & 27.6 & ~ & 43.0 & ~ & 29.3 & ~\\
MixFormer-B4(\textbf{Ours}) & \textbf{28.6} & \cgaphlp{+}{1.0} & \textbf{43.4} & \cgaphlp{+}{0.4} & \textbf{30.5} & \cgaphlp{+}{1.2}\\
\shline
\end{tabular}
\caption{\textbf{More Downstream Tasks.} We compare our MixFormer with ResNet50~\cite{he2016deep} and Swin Transformer~\cite{liu2021swin} on keypoint detection and long-tail instance segmentation.}
\label{tab:12}\vspace{-2mm}
\end{table}
%##################################################################################################

\subsection{Generalization}
\paragraph{More Downstream Tasks.} In Table~\ref{tab:12}, we conduct experiments on two more downstream tasks: keypoint detection and long-tail instance segmentation. Detailed experimental settings are provided in Appendix~\ref{Appendix:C}.

{\em COCO keypoint Detection}: In Table~\ref{tab:12}, MixFormer-B4 outperforms baseline models~\cite{he2016deep,liu2021swin} by significant margins in all metrics. Moreover, MixFormer also shows clear advantages compared with HRFormer~\cite{YuanFHLZCW21}, which is specifically designed for dense prediction tasks.

{\em LVIS $1.0$ Instance Segmentation:} This task has $\sim1000$ long-tailed distribution categories, which relies on the discriminative feature learned by the backbone. Results in Table~\ref{tab:12} show that MixFormer outperforms the Swin-T~\cite{liu2021swin} by $1.0$ AP$^\text{mask}$, which demonstrates the robustness of the learned representation in MixFormer.
%##################################################################################################
\begin{table}
\small
\centering
\addtolength{\tabcolsep}{-4.0pt}
\begin{tabular}{c|c|cl|cl}
\shline
Models & FLOPs& \multicolumn{2}{l|}{Top-1} & \multicolumn{2}{l}{Top-5}\\
\hline
ResNet50~\cite{touvron2021training} & 4.1G&  78.4 & ~ & - & ~\\
ResNet50~\cite{wightman2021resnet} & 4.1G&  79.8 & ~ & - & ~\\
ResNet50$^*$ & 4.1G& 79.0 & ~ & 94.3 & ~\\
ResNet50 + Mixing Block & 3.9G& \textbf{80.6} & \cgaphlp{+}{1.6} & \textbf{95.1} & \cgaphlp{+}{0.8}\\
\hline
MobileNetV2~\cite{sandler2018mobilenetv2} & 0.3G& 72.0 & ~ & - & ~\\
MobileNetV2$^*$ & 0.3G& 71.7 & ~ & 90.3 & ~\\
MobileNetV2+SE+Non-Local$^*$ & 0.3G& 72.5 & ~ & 91.0 & ~\\
MobileNetV2 + Mixing Block & 0.3G& \textbf{73.6} & \cgaphlp{+}{1.9} & \textbf{91.6} & \cgaphlp{+}{1.3}\\
\shline
\end{tabular}
\caption{\textbf{Apply Mixing Block to ConvNets on ImageNet-1K.} We introduce our Mixing Block to typical ConvNets, ResNet~\cite{he2016deep} and MobileNetV2~\cite{sandler2018mobilenetv2}. As different training recipes give variant accuracy~\cite{wightman2021resnet}, we also train ResNet50~\cite{he2016deep} and MobileNetV2~\cite{sandler2018mobilenetv2} with the same setting as ours, denoted with $*$ in the Table.}
\label{tab:13}\vspace{-4mm}
\end{table}
%##################################################################################################

{\em Summary:} Considering the promising results given by MixFormer in previous tasks: object detection, instance segmentation, and semantic segmentation, MixFomer can serve as a general-purpose backbone and outperform its alternatives in $\textbf{5}$ dense prediction tasks.

\paragraph{Apply Mixing Block to ConvNets.} We apply our Mixing Block to typical ConvNets, ResNet50~\cite{he2016deep} and MobileNetV2~\cite{sandler2018mobilenetv2}. Following~\cite{srinivas2021bottleneck}, we replace all the blocks in the last stage with our Mixing Block in ConvNets. To make a fair comparison, we adjust the number of blocks to maintain the overall computational cost. Table~\ref{tab:13} shows that the Mixing Block can provide gains on ConvNets~\cite{he2016deep,sandler2018mobilenetv2} as a alternative to ConvNet blocks. Specifically, Mixing Block brings $1.9\%$ and $1.6\%$ Top-1 accuracy on ImageNet-1K~\cite{deng2009imagenet} over MobileNetV2~\cite{sandler2018mobilenetv2} and ResNet50~\cite{he2016deep}. Moreover, we also provide the result of MobileNetV2~\cite{sandler2018mobilenetv2} with SE layer~\cite{hu2018squeeze} and Non-Local~\cite{wang2018non} in Table~\ref{tab:13}. It gives inferiror performance than our mixing block.

%------------------------------------------------------------------------
\section{Limitations}
\label{sec:limits}
%-------------------------------------------------------------------------
Our MixFormer is proposed to mitigate the issues in local-window self-attention~\cite{vaswani2021scaling,liu2021swin}. Thus it may be limited to window-based vision transformers in this paper. Although the parallel design and the bi-directional interactions can be applied to the global self-attention~\cite{dosovitskiy2020image,touvron2021training}, it is not clear that how many gains can the above designs bring. We conduct a simple experiment on DeiT-Tiny~\cite{touvron2021training}. But the result becomes slightly worse, as shown in Table~\ref{tab:B2}. More efforts are needed to apply our mixing block to global attention. We leave this for future work. Moreover, we build the MixFormer series manually, limiting MixFormer in existing instances. Other methods such as NAS (Network Architecture Search)~\cite{tan2019efficientnet} can be applied to further improve the results.

%------------------------------------------------------------------------
\section{Conclusion}
\label{sec:conclu}
%-------------------------------------------------------------------------
In this paper, we propose MixFormer as an efficient general-purpose vision transformer. Addressing issues in Window-based Vision Transformer, we seek to alleviate limited receptive fields and weak modeling capability on the channel dimension. Our MixFormer enlarges receptive fields efficiently without shifting or shuffling windows, thanks to a parallel design coupling local window and depth-wise convolution. The bi-directional interactions boost modeling ability in the channel and spatial dimension for local-window self-attention and depth-wise convolution, respectively. Extensive experiments show that MixFormer outperforms its alternatives on image classification and various downstream vision tasks. We expect the designs in MixFormer to serve as a base setup for designing efficient networks. 
%-------------------------------------------------------------------------
\section*{Acknowledgements}
The authors would like to thank Jiaying Guo, Zhe Li, and Fanrong Li for their helpful discussions and feedback. This work was supported in part by National Natural Science Foundation of China (No.62106267)
%%%%%%%%%%%%%%%%%%%%%%%%%%%%%%%%%%%%%%%%%%%%%%%%%%%%%%%%%%%%%%%%%%%%%%%%%%%%%%%%%%%%%%%%%%%%%%%%%%%
\appendix
\renewcommand{\thesection}{\Alph{section}}
\section*{Appendix}
%##################################################################################################
\begin{table}
\tablestyle{1.pt}{1.1}
\begin{tabular}{x{65}x{35}x{50}x{55}}
\shline
Models & \#Channels   & \#Blocks & \#Heads\\
\hline
MixFormer-B0   &  $C=24$ & $[1, 2, 6, 6]$ & $[3, 6, 12, 24]$\\
MixFormer-B1   &  $C=32$ & $[1, 2, 6, 6]$ & $[2, 4, 8, 16]$\\
MixFormer-B2   &  $C=32$ & $[2, 2, 8, 8]$ & $[2, 4, 8, 16]$\\
MixFormer-B3   &  $C=48$ & $[2, 2, 8, 6]$ & $[3, 6, 12, 24]$\\
MixFormer-B4   &  $C=64$ & $[2, 2, 8, 8]$ & $[4, 8, 16, 32]$\\
MixFormer-B5   &  $C=96$ & $[1, 2, 8, 6]$ & $[6, 12, 24, 48]$\\
MixFormer-B6   &  $C=96$ & $[2, 4, 16, 12]$ & $[6, 12, 24, 48]$\\
\shline
\end{tabular}
\caption{\textbf{Architecture Variants.} Detailed configurations of architecture variants of MixFormer.}
\label{tab:A1}\vspace{-5mm}
\end{table}
%##################################################################################################
\section{More Variants of MixFormer} \label{Appendix:A}
We scale our MixFormer to smaller and larger models. In this section, we provide two instantiated models (MixFormer-B0 and MixFormer-B5). Their detailed settings are provided in Table~\ref{tab:A1}, along with previous methods (from B1 to B4). Note that MixFormer-B0 and MixFormer-B5 are two examples. More variants can be obtained with further attempts following the design of MixFormer. Then, we validate their effectiveness on ImageNet-1K~\cite{deng2009imagenet}. The results are illustrated in Table~\ref{tab:A2}.

On one side, MixFormer-B0 achieves competitive result ($76.5\%$ Top-1 accuracy on ImageNet-1K~\cite{deng2009imagenet}) even with $0.4$G FLOPs, which lies in the mobile level~\cite{sandler2018mobilenetv2,ma2018shufflenet}. While other vision transformer variants~\cite{wang2021pyramid,wu2021cvt,wang2021pvtv2,liu2021swin,yang2021focal} did not provide a range of model sizes like our MixFormer, especially in mobile level. We believe that further efforts can be made to give higher performance to achieve state-of-the-art results~\cite{tan2019efficientnet,howard2019searching} in mobile level models. On the other side, MixFormer-B5 shows an example to scale our MixFormer to larger models. It has $6.8$G FLOPs, while it can achieve on par results with Swin-B (15.4G)~\cite{liu2021swin}, Focal-S (9.1G)~\cite{yang2021focal}, Shuffle-S (8.9G)~\cite{huang2021shuffle}, and EfficientNet-B5 (9.9G)~\cite{tan2019efficientnet}, which demonstrates the computational efficiency of MixFormer. MixFormer-B6 achieves \textbf{83.8\%} top-1 accuracy on ImageNet-1K~\cite{deng2009imagenet}. It maintains the superior performance to Swin-B(15.4G)~\cite{liu2021swin} and is comparable to other models with less flops.

The above results verify the scalability of MixFormer to smaller and larger models. Moreover, it has the potential for further improvements.
%##################################################################################################
\begin{table}
\centering
\tablestyle{5.pt}{1.05}
\begin{tabular}{cccc}
\shline
\multicolumn{1}{c|}{Method} &  \multicolumn{1}{l|}{\#Params} & \multicolumn{1}{l|}{FLOPs} & \multicolumn{1}{l}{Top-1} \\ 
\midrule
\multicolumn{4}{c}{ConvNets} \\  \midrule
\multicolumn{1}{c|}{RegNetY-4G~\cite{radosavovic2020designing}}
& \multicolumn{1}{c|}{21M} & \multicolumn{1}{c|}{4.0G} & \multicolumn{1}{c}{80.0}\\ 
\multicolumn{1}{c|}{RegNetY-8G~\cite{radosavovic2020designing}}
& \multicolumn{1}{c|}{39M} & \multicolumn{1}{c|}{8.0G}  & \multicolumn{1}{c}{81.7}\\
\multicolumn{1}{c|}{RegNetY-16G~\cite{radosavovic2020designing}}
& \multicolumn{1}{c|}{84M} & \multicolumn{1}{c|}{16.0G}  & \multicolumn{1}{c}{82.9}\\
\multicolumn{1}{c|}{EffNet-B0~\cite{tan2019efficientnet}}
& \multicolumn{1}{c|}{5M} & \multicolumn{1}{c|}{0.4G} & \multicolumn{1}{c}{77.1}\\
\multicolumn{1}{c|}{EffNet-B1~\cite{tan2019efficientnet}}
& \multicolumn{1}{c|}{8M} & \multicolumn{1}{c|}{0.7G} & \multicolumn{1}{c}{79.1}\\
\multicolumn{1}{c|}{EffNet-B2~\cite{tan2019efficientnet}}
& \multicolumn{1}{c|}{9M} & \multicolumn{1}{c|}{1.0G} & \multicolumn{1}{c}{80.1}\\
\multicolumn{1}{c|}{EffNet-B3~\cite{tan2019efficientnet}}
& \multicolumn{1}{c|}{12M} & \multicolumn{1}{c|}{1.8G} & \multicolumn{1}{c}{81.6}\\
\multicolumn{1}{c|}{EffNet-B4~\cite{tan2019efficientnet}}
& \multicolumn{1}{c|}{19M} & \multicolumn{1}{c|}{4.2G} & \multicolumn{1}{c}{82.9}\\
\multicolumn{1}{c|}{EffNet-B5~\cite{tan2019efficientnet}}
& \multicolumn{1}{c|}{30M} & \multicolumn{1}{c|}{9.9G} & \multicolumn{1}{c}{83.6}\\
\hline
\multicolumn{4}{c}{Vision Transformers}\\
\hline
\multicolumn{1}{c|}{DeiT-T~\cite{touvron2021training}}& \multicolumn{1}{c|}{6M}
& \multicolumn{1}{c|}{1.3G} & \multicolumn{1}{c}{72.2} \\
\multicolumn{1}{c|}{DeiT-S~\cite{touvron2021training}}
&\multicolumn{1}{c|}{22M} & \multicolumn{1}{c|}{4.6G}
& \multicolumn{1}{c}{79.9}\\
\multicolumn{1}{c|}{DeiT-B~\cite{touvron2021training}}
&\multicolumn{1}{c|}{87M} & \multicolumn{1}{c|}{17.5G}
& \multicolumn{1}{c}{81.8}\\
\multicolumn{1}{c|}{PVT-T~\cite{wang2021pyramid}} & \multicolumn{1}{c|}{13M}  
& \multicolumn{1}{c|}{1.8G} &\multicolumn{1}{c}{75.1} \\
\multicolumn{1}{c|}{PVT-S~\cite{wang2021pyramid}}
& \multicolumn{1}{c|}{25M} & \multicolumn{1}{c|}{3.8G}
& \multicolumn{1}{c}{79.8}\\
\multicolumn{1}{c|}{PVT-M~\cite{wang2021pyramid}}& \multicolumn{1}{c|}{44M} & \multicolumn{1}{c|}{6.7G}
& \multicolumn{1}{c}{81.2}\\
\multicolumn{1}{c|}{PVT-L~\cite{wang2021pyramid}}& \multicolumn{1}{c|}{61M}
& \multicolumn{1}{c|}{9.8G} & \multicolumn{1}{c}{81.7}\\
\multicolumn{1}{c|}{CvT-13\cite{wu2021cvt}}& \multicolumn{1}{c|}{20M}
& \multicolumn{1}{c|}{4.5G} & \multicolumn{1}{c}{81.6}\\
\multicolumn{1}{c|}{CvT-21\cite{wu2021cvt}} & \multicolumn{1}{c|}{32M}
& \multicolumn{1}{c|}{7.1G} & \multicolumn{1}{c}{82.5}\\
\multicolumn{1}{c|}{TwinsP-S~\cite{chu2021twins}}  & \multicolumn{1}{c|}{24M}
& \multicolumn{1}{c|}{3.8G} & \multicolumn{1}{c}{81.2}\\
\multicolumn{1}{c|}{DS-Net-S\cite{mao2021dual}} & \multicolumn{1}{c|}{23M}
& \multicolumn{1}{c|}{3.5G} & \multicolumn{1}{c}{82.3}\\
\hline
\multicolumn{1}{c|}{Swin-T~\cite{liu2021swin}}& \multicolumn{1}{c|}{29M}
& \multicolumn{1}{c|}{4.5G} & \multicolumn{1}{c}{81.3}\\
\multicolumn{1}{c|}{Swin-S~\cite{liu2021swin}}& \multicolumn{1}{c|}{50M}
& \multicolumn{1}{c|}{8.7G} & \multicolumn{1}{c}{83.0}\\
\multicolumn{1}{c|}{Swin-B~\cite{liu2021swin}}& \multicolumn{1}{c|}{88M}
& \multicolumn{1}{c|}{15.4G} & \multicolumn{1}{c}{83.5}\\
\multicolumn{1}{c|}{Twins-S~\cite{chu2021twins}}& \multicolumn{1}{c|}{24M}
& \multicolumn{1}{c|}{2.9G} & \multicolumn{1}{c}{81.7}\\
\multicolumn{1}{c|}{Twins-B~\cite{chu2021twins}}& \multicolumn{1}{c|}{56M}
& \multicolumn{1}{c|}{8.6G} & \multicolumn{1}{c}{83.2}\\
\multicolumn{1}{c|}{LG-T\cite{li2021local}}& \multicolumn{1}{c|}{33M}
& \multicolumn{1}{c|}{4.8G} & \multicolumn{1}{c}{82.1}\\
\multicolumn{1}{c|}{LG-S\cite{li2021local}} & \multicolumn{1}{c|}{61M}
& \multicolumn{1}{c|}{9.4G} & \multicolumn{1}{c}{83.3}\\
\multicolumn{1}{c|}{Focal-T\cite{yang2021focal}}& \multicolumn{1}{c|}{29M}       & \multicolumn{1}{c|}{4.9G} & \multicolumn{1}{c}{82.2}\\
\multicolumn{1}{c|}{Focal-S\cite{yang2021focal}}& \multicolumn{1}{c|}{51M}       & \multicolumn{1}{c|}{9.1G} & \multicolumn{1}{c}{83.5}\\
\multicolumn{1}{c|}{Shuffle-T\cite{huang2021shuffle}}& \multicolumn{1}{c|}{29M}       & \multicolumn{1}{c|}{4.6G} & \multicolumn{1}{c}{82.5}\\
\multicolumn{1}{c|}{Shuffle-S\cite{huang2021shuffle}}& \multicolumn{1}{c|}{50M}       & \multicolumn{1}{c|}{8.9G} & \multicolumn{1}{c}{83.5}\\
\hline
\multicolumn{1}{c|}{MixFormer-B0 (\textbf{Ours})}  & \multicolumn{1}{c|}{5M}  & \multicolumn{1}{c|}{0.4G}  &\multicolumn{1}{c}{76.5}\\
\multicolumn{1}{c|}{MixFormer-B1 (\textbf{Ours})}  & \multicolumn{1}{c|}{8M}  & \multicolumn{1}{c|}{0.7G}  &\multicolumn{1}{c}{78.9}\\
\multicolumn{1}{c|}{MixFormer-B2 (\textbf{Ours})}
& \multicolumn{1}{c|}{10M}  & \multicolumn{1}{c|}{0.9G}
& \multicolumn{1}{c}{80.0}\\
\multicolumn{1}{c|}{MixFormer-B3 (\textbf{Ours})}  & \multicolumn{1}{c|}{17M} & \multicolumn{1}{c|}{1.9G} 
& \multicolumn{1}{c}{81.7}\\
\multicolumn{1}{c|}{MixFormer-B4 (\textbf{Ours})}  & \multicolumn{1}{c|}{35M}  & \multicolumn{1}{c|}{3.6G}  & \multicolumn{1}{c}{83.0}\\
\multicolumn{1}{c|}{MixFormer-B5 (\textbf{Ours})}  & \multicolumn{1}{c|}{62M}  & \multicolumn{1}{c|}{6.8G}  & \multicolumn{1}{c}{83.5}\\
\multicolumn{1}{c|}{MixFormer-B6 (\textbf{Ours})}  & \multicolumn{1}{c|}{119M} & \multicolumn{1}{c|}{12.7G}  & \multicolumn{1}{c}{83.8}\\
\shline
\end{tabular}
\caption{\textbf{Classification accuracy on the ImageNet validation set.} Performances are measured with a single $224\times224$ crop. ``Params'' refers to the number of parameters. ``FLOPs'' is calculated under the input scale of $224\times 224$.}
\label{tab:A2}\vspace{-5mm}
\end{table}
%##################################################################################################
\section{Additional Experiments} \label{Appendix:B}
\paragraph{Window Sizes in Local-Window Self-Attention.} We conduct ablation study on the window size in local-window self-attention with MixFormer-B1. The experimental settings are follow the ones in the ablation studies. The results in Table~\ref{tab:B1} show that larger window size (ws=12) achieves on par performance with ws=7 (78.4\%) on ImageNet-1K~\cite{deng2009imagenet}. Based on the above result, We follow the con-
ventional design of Swin Transformer (ws=7)~\cite{liu2021swin} in all variants of MixFormer.
%##################################################################################################
\begin{table}
\small
\centering
\addtolength{\tabcolsep}{-2.5pt}
\begin{tabular}{c|cc}
\shline
\multirow{2}{*}{Window Sizes} & \multicolumn{2}{c}{ImageNet} \\
 & Top-1 & Top-5\\
\hline
$7\times7$ & 78.4 & 94.3 \\
$12\times12$ & 78.4 & 94.5 \\
\shline
\end{tabular}
\caption{\textbf{Window Sizes in Local-window Self-attention.} We investigate various window sizes for Local-window Self-attention in MixFormer.}
\label{tab:B1}\vspace{-2mm}
\end{table}
%##################################################################################################
%##################################################################################################
\begin{table}
\small
\centering
\addtolength{\tabcolsep}{-2.5pt}
\begin{tabular}{c|cc}
\shline
\multirow{2}{*}{DeiT-Tiny~\cite{touvron2021training}} & \multicolumn{2}{c}{ImageNet} \\
 & Top-1 & Top-5\\
\hline
Baseline & 72.2 & 91.1 \\
Baseline+Mixing Block & 71.3 & 90.5 \\
\shline
\end{tabular}
\caption{\textbf{Apply Mixing Block to DeiT-Tiny.} We apply our mixing block to global attention.}
\label{tab:B2}\vspace{-2mm}
\end{table}
%##################################################################################################
\paragraph{Apply Mixing Block to DeiT.} Although our mixing block is proposed to solve the window connection problem in local-window self-attention~\cite{liu2021swin}. It can also be applied to global attentions~\cite{dosovitskiy2020image,touvron2021training}. We simply apply our mixing block to Deit-Tiny~\cite{touvron2021training}. But the result is slightly lower than baseline (71.3\% vs. 72.2\%) on ImageNet-1K~\cite{deng2009imagenet}. We conjecture that global attention (ViT-based model) may not share the same problem and detailed design for global attention may need further investigating. We leave this for future work.

\section{Detailed Experimental Settings} \label{Appendix:C}
%##################################################################################################
\begin{figure}
\centering
\includegraphics[width=0.42\textwidth]{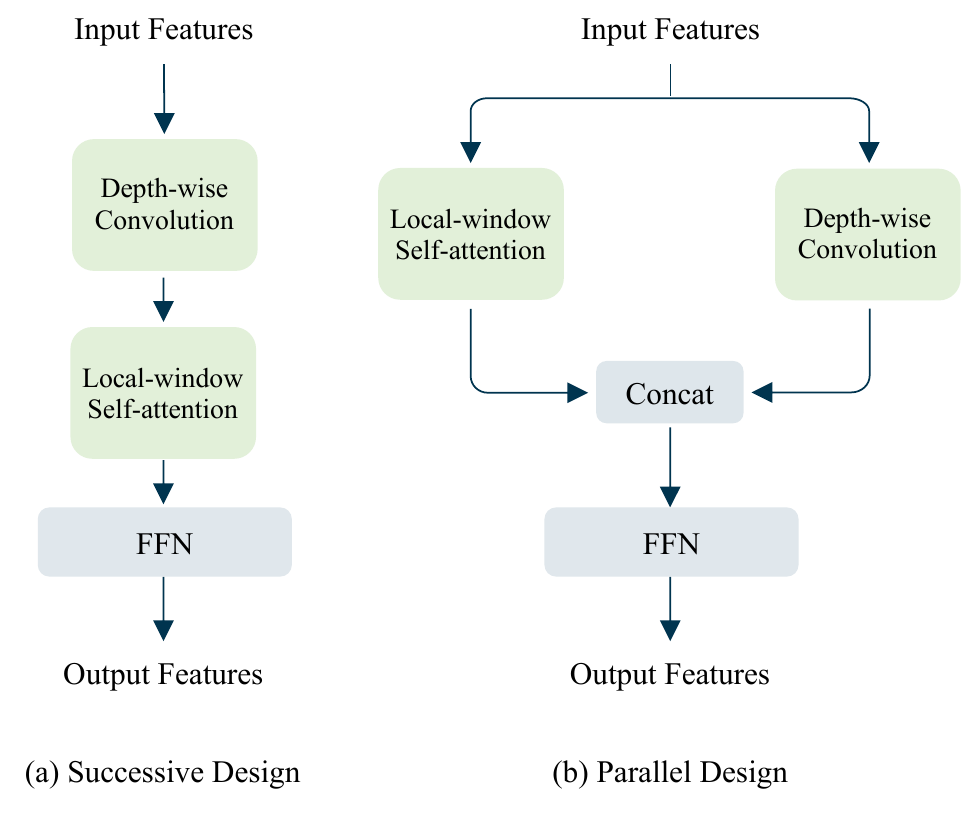}
\caption{\textbf{Successive Design and Parallel Design.} We combine local-window self-attention with depth-wise convolution in two different ways. Other details in the block, such as module design, normalization layers, and shortcuts, are omitted for a neat presentation.}
\label{fig:B1}\vspace{-1mm}
\end{figure}
%##################################################################################################
\paragraph{Successive Design and Parallel Design.} In Figure~\ref{fig:B1}, we give the details on how to combine local-window self-attention and depth-wise convolution in the successive design and the parallel design. To make a fair comparison, we adjust the channels in the blocks to keep the computational complexity the same in the two designs.

\paragraph{Image Classification.} We train all models for 300 epochs with an image size of $224\times224$ on ImageNet-1K~\cite{deng2009imagenet}. We adjust the training settings gently when training models in different sizes. The detailed setting is in Table~\ref{tab:A3}.
%##################################################################################################
\begin{table}[t]
\tablestyle{6.5pt}{1.}
\scriptsize
\begin{tabular}{l|l}
\shline
config & value \\
\hline
optimizer & AdamW~\cite{2017Fixing} \\
base learning rate & 8e-4 (B0-B3), 1e-3 (B4, B5, B6) \\
weight decay & 0.04 (B0-B3), 0.05 (B4, B5, B6) \\
optimizer momentum & $\beta_1, \beta_2{=}0.9, 0.999$ \\
batch size & 1024 \\
learning rate schedule & cosine decay~\cite{loshchilov2016sgdr} \\
minimum learning rate & 1e-6 \\
warmup epochs & 20 (B0-B4), 40 (B5, B6) \\
warmup learning rate & 1e-7 \\
training epochs & 300  \\
augmentation & RandAug(9, 0.5)~\cite{cubuk2020randaugment} \\
color jitter & 0.4 \\
mixup~\cite{zhang2017mixup} & 0.2 \\
cutmix~\cite{yun2019cutmix} & 1.0 \\
random erasing~\cite{zhong2020random} & 0.25 \\ 
drop path~\cite{huang2016deep} & [0.0, 0.05, 0.1, 0.2, 0.3, 0.5] (B0-B6) \\
% exp. moving average (EMA) & 0.9999
\shline
\end{tabular}
\caption{\textbf{Image Classification Training Settings.}}
\label{tab:A3}
\vspace{-1em}
\end{table}
%##################################################################################################

%##################################################################################################
\begin{table}[t]
\tablestyle{7pt}{1.}
\scriptsize
\begin{tabular}{l|l}
\shline
config & value \\
\hline
optimizer & AdamW \\
base learning rate & 0.0002 \\
weight decay & 0.04 (B0-B3), 0.05 (B4, B5) \\
optimizer momentum & $\beta_1, \beta_2{=}0.9, 0.999$ \\
batch size & 16 \\
learning rate schedule & steps:[8, 11] ($1\times$), [27, 33] ($3\times$) \\
warmup iterations (ratio) & 500 (0.001) \\
training epochs & 12 ($1\times$), 36 ($3\times$)  \\
scales & (800, 1333) ($1\times$), Multi-scales~\cite{liu2021swin} ($3\times$) \\
drop path & 0.0 (B0-B3), 0.1 (B4, B5) \\
% exp. moving average (EMA) & 0.9999
\shline
\end{tabular}
\caption{\textbf{Object Detection and Instance Segmentation Training Settings.}}
\label{tab:A4}
\vspace{-2em}
\end{table}
%##################################################################################################
\paragraph{Object Detection and Instance Segmentation.} When transferring MixFormer to object detection and instance segmentation on MS COCO~\cite{lin2014microsoft}, we consider two typical frameworks: Mask R-CNN~\cite{he2017mask} and Cascade Mask R-CNN~\cite{he2017mask,cai2018cascade}. We adopt AdamW~\cite{2017Fixing} optimizer with an initial learning rate of $0.0002$ and a batch size of $16$. To make a fair comparison with other works, we make all normalization layers trainable in MixFormer\footnote{Wherever BN is applied, we use {\em synchronous} BN across all GPUs.}. When training different sizes of models, we adjust the training settings gently according to their settings used in image classification. Table~\ref{tab:A4} shows the detailed hyper-parameters used in training models on MS COCO~\cite{lin2014microsoft}.

\paragraph{Semantic Segmentation.} On ADE20K~\cite{zhou2019semantic}, we use the AdamW optimizer~\cite{2017Fixing} with an initial learning rate $0.00006$, a weight decay $0.01$, and a batch size of $16$. We train all models for $160$K on ADE20K. For testing, we report the results with single-scale testing and multi-scale testing on main comparisons, while we only give single-scale testing results on ablation studies. In multi-scale testing, the resolutions used are the $[0.5, 0.75, 1.0, 1.25, 1.5, 1.75]\times$ of that in training. The settings mainly follow~\cite{liu2021swin}. For the path drop rates in different models, we adopt the same hyper-parameters as in MS COCO~\cite{lin2014microsoft}.

\paragraph{Keypoint Detection.} We conduct experiments on the MS COCO human pose estimation benchmark. We train the models for $210$ epochs with an AdamW optimizer, an image size of $256\times192$, and a batch size of $256$. The training and evaluation hyper-parameters are mostly following the ones in HRFormer~\cite{YuanFHLZCW21}.

\paragraph{Long-tail Instance Segmentation.} We use the hyper-parameters of Mask R-CNN~\cite{he2017mask} on MS COCO~\cite{lin2014microsoft} when training models for long-tail instance segmentation on LVIS~\cite{gupta2019lvis}. We report the results with a $1\times$ schedule. The training augmentations and sampling methods are the same for all models, which adopt a multi-scale training and use balanced sampling by following~\cite{gupta2019lvis}.

\section{Discussion with Related Works} \label{Appendix:D}
In MixFormer, we consider two types of information exchanges: (1) across dimensions, (2) across windows. 

For the first type, Conformer~\cite{peng2021conformer} also performs information exchange between a transformer branch and a convolution branch. While its motivation is different from ours. Conformer aims to couple local and global features across convolution and transformer branches. MixFormer uses channel and spatial interactions to address the weak modeling ability issues caused by weight sharing on the channel (local-window self-attention) and the spatial (depth-wise convolution) dimensions~\cite{han2021demystifying}. 

For the second type, Twins (strided convolution + global sub-sampled attention)~\cite{chu2021twins} and Shuffle Transformer (neighbor-window connection (NWC) + random spatial shuffle)~\cite{huang2021shuffle} construct local and global connections to achieve information exchanges, MSG Transformer (channel shuffle on extra MSG tokens)~\cite{fang2021msg} applies global connection. Our MixFormer achieves this goal by concatenating the parallel features: the non-overlapped window feature and the local-connected feature (output of the dwconv3x3).

%%%%%%%%% REFERENCES
{\small
\bibliographystyle{ieee_fullname}
\bibliography{main}
}
\end{document}